%% file: aaai23.tex
\documentclass[letterpaper]{article} 
\usepackage{aaai23}  
\usepackage{times}  
\usepackage{helvet}  
\usepackage{courier}  
\usepackage[hyphens]{url}  
\usepackage{graphicx} 
\usepackage{natbib}  
\usepackage{caption} 
\usepackage{newfloat}
\usepackage{listings}
\usepackage{amsmath}
\usepackage{amsfonts}
\usepackage{graphicx}
\usepackage{subcaption}
\usepackage{amssymb}
\usepackage[ruled]{algorithm2e}
\usepackage{bm}	
\usepackage{mathtools}
\usepackage{newfloat}
\usepackage{listings}
\usepackage{newfloat}
\usepackage{listings}
\usepackage{url}

\newcommand{\bx}{\boldsymbol{x}}

\newcommand{\cB}{\mathcal{B}}
\newcommand{\bI}{\boldsymbol{I}}

\newcommand{\norm}[1]{\left\| #1 \right\|}

\newcommand{\cA}{\mathcal{A}}

\newcommand{\cK}{\mathcal{K}}

\newtheorem{theorem}{Theorem}

\newtheorem{lemma}[theorem]{Lemma}

\newtheorem{definition}{\bf Definition}
\newenvironment{proof}{\noindent {\textbf{Proof. }}}{$\Box$ \medskip}
\DeclareCaptionStyle{ruled}{labelfont=normalfont,labelsep=colon,strut=off} 
\title{Efficient Explorative Key-term Selection Strategies for Conversational Contextual
Bandits}
\author{
    Zhiyong Wang,\textsuperscript{\rm 1}
    Xutong Liu,\textsuperscript{\rm 1}
    Shuai Li,\textsuperscript{\rm 2}\thanks{Corresponding author.}
    John C.S. Lui\textsuperscript{\rm 1}
}
\affiliations{
    \textsuperscript{\rm 1}The Chinese University of Hong Kong
    ,Hong Kong SAR, China

    \textsuperscript{\rm 2}Shanghai Jiao Tong University,
    Shanghai, China\\
    	zywang21@cse.cuhk.edu.hk, liuxt@cse.cuhk.edu.hk, shuaili8@sjtu.edu.cn, 	cslui@cse.cuhk.edu.hk
}

\usepackage{bibentry}

\begin{document}
\maketitle
\begin{abstract}
Conversational contextual bandits elicit user preferences by occasionally querying for explicit feedback on key-terms to accelerate learning. 
However, there are aspects of existing approaches which limit their performance.
First, information gained from key-term-level conversations and arm-level recommendations is not appropriately incorporated to speed up learning.
Second, it is important to ask explorative key-terms to quickly elicit the user's potential interests in various domains to accelerate the convergence of user preference estimation, which has never been considered in existing works. 
To tackle these issues, we first propose ``ConLinUCB", a general framework for conversational bandits with better information incorporation, combining arm-level and key-term-level feedback to estimate user preference in one step at each time. Based on this framework, we further design two bandit algorithms with explorative key-term selection strategies, ConLinUCB-BS and ConLinUCB-MCR. We prove tighter regret upper bounds of our proposed algorithms. Particularly, ConLinUCB-BS achieves a regret bound of $O(d\sqrt{T\log T})$, better than the previous result $O(d\sqrt{T}\log T)$. Extensive experiments on synthetic and real-world data show significant advantages of our algorithms in learning accuracy (up to 54\% improvement) and computational efficiency (up to 72\% improvement), compared to the classic ConUCB algorithm, showing the potential benefit to recommender systems. 
\end{abstract}

\section{Introduction}
Nowadays, recommender systems are widely used in various areas. The learning speed for traditional online recommender systems is usually slow since extensive exploration is needed to discover user preferences. To accelerate the learning process and provide more personalized recommendations, the conversational recommender system (CRS) has been proposed \cite{christakopoulou2016towards,christakopoulou2018q,sun2018conversational,zhang2018towards,li2021seamlessly,gao2021advances}. In a CRS,  a learning agent occasionally asks for the user's explicit feedback on some ``key-terms", and leverages this additional conversational information to better elicit the user's preferences \cite{zhang2020conversational,xie2021comparison}. 
    
Despite the recent success of CRS, there are crucial limitations in using conversational contextual bandit approaches to design recommender systems. These limitations include: (a) The information gained from key-term-level conversations and arm-level recommendations is not incorporated properly to speed up learning, as the user preferences are essentially assumed to be the same in these two stages but are estimated separately \cite{zhang2020conversational,xie2021comparison,wu2021clustering}; 
(b) Queries using traditional key-terms were restrictive and not explorative enough. Specifically, we say a key-term is ``explorative" if it is under-explored so far and the system is uncertain about the user's preferences in its associated items. Asking for the user's feedback on explorative key-terms can efficiently elicit her potential interests in various domains (e.g., sports, science), which means we can quickly estimate the user preference vector in all directions of the feature space, thus accelerating the learning speed. Therefore, it is crucial to design explorative key-term selection strategies, which existing works have not considered.
    
Motivated by the above considerations, we propose to design conversational bandit algorithms that (i) estimate the user's preferences utilizing both arm-level and key-term-level interactions simultaneously to properly incorporate the information gained from both two levels and (ii) use effective strategies to choose explorative key-terms when conducting conversations for quick user preference inference.
    
To better utilize the interactive feedback from both recommendations and conversations, we propose ConLinUCB, a general framework for conversational bandits with possible flexible key-term selection strategies. ConLinUCB estimates the user preference vector by solving \textit{one single optimization problem} that minimizes the mean squared error of both arm-level estimated rewards and key-term-level estimated feedback simultaneously, instead of separately estimating at different levels as in previous works. In this manner, the information gathered from these two levels can be better combined to guide the learning. 

Based on this ConLinUCB framework, we design two new algorithms with explorative key-term selection strategies, ConLinUCB-BS and ConLinUCB-MCR.
\begin{itemize}
    \item ConLinUCB-BS makes use of a barycentric spanner containing linearly independent vectors, which can be an efficient exploration basis in bandit problems \cite{amballa2021computing}. Whenever a conversation is allowed, ConLinUCB-BS selects an explorative key-term uniformly at random from a precomputed barycentric spanner $\mathcal{B}$ of the given key-term set $\mathcal{K}$. 
    \item
    ConLinUCB-MCR applies in a more general setting when the key-term set can be time-varying, and it can leverage interactive histories to choose explorative key-terms adaptively. Note that in the bandit setting, we often use confidence radius to adaptively evaluate whether an arm has been sufficiently explored, and the confidence radius of an arm will shrink whenever it is selected  \cite{lattimore2020bandit}. This implies that an explorative key-term should have a large confidence radius. Based on this reasoning, ConLinUCB-MCR selects the most explorative key-terms with maximal confidence radius when conducting conversations. 
\end{itemize}

Equipped with explorative conversations, our algorithms can quickly elicit user preferences for better recommendations. For example, if the key-term \textit{sports} is explorative at round $t$, indicating that so far the agent is not sure whether the user favors items associated with \textit{sports} (e.g., basketball, volleyball), it will ask for the user's feedback on \textit{sports} directly and conduct recommendations
accordingly. In this manner, the agent can quickly find suitable items for the user. We prove the regret upper bounds of our algorithms, which are better than the classic ConUCB algorithm.
    
In summary, our paper makes the following contributions:
\begin{itemize}
    \item We propose a new and general framework for conversational contextual bandits, ConLinUCB, which can efficiently incorporate the interactive information gained from both recommendations and conversations.
    \item  Based on ConLinUCB, we design two new algorithms with explorative key-term selection strategies, ConLinUCB-BS and ConUCB-MCR, which can accelerate the convergence of user preference estimation. 
    \item  We prove that our algorithms achieve tight regret upper bounds. Particularly, ConLinUCB-BS achieves a bound of $O(d\sqrt{T\log T})$, better than the previous $O(d\sqrt{T}\log T)$ in the conversational bandits literature.
    \item Experiments on both synthetic and real-world data validate the advantages of our algorithms in both learning accuracy (up to 54\% improvement) and computational efficiency (up to 72\% improvement)\footnote{Codes are available at \url{https://github.com/ZhiyongWangWzy/ConLinUCB.}}.
\end{itemize}

\section{Related Work} \label{section7}

Our work is most closely related to the research on conversational contextual bandits. 

Contextual linear bandit is an online sequential decision-making problem where at each time step, the agent has to choose an action and receives a corresponding reward whose expected value is an unknown linear function of the action \cite{li2010contextual,chu2011contextual,
abbasi2011improved,wu2016contextual}. The objective is to collect as much reward as possible in $T$ rounds. 

Traditional linear bandits need extensive exploration to capture the user preferences in recommender systems. To speed up online recommendations, the idea of conversational contextual bandits was first proposed in \cite{zhang2020conversational}, where 
conversational feedback on key-terms is leveraged to assist the user preference elicitation. In that work, they propose the ConUCB algorithm with a theoretical regret bound of $O(d\sqrt{T}\log T)$. Some follow-up works try to improve the performance of ConUCB with the help of additional information, such as self-generated key-terms~\cite{wu2021clustering}, relative feedback~\cite{xie2021comparison}, and knowledge graph~\cite{zhao2022knowledge}. 
Unlike these works, we adopt the same problem settings as ConUCB and improve the underlying mechanisms without relying on additional information.
Yet one can use the principles of efficient information incorporation and explorative conversations proposed in this work to enhance these works when additional information is available, which is left as an interesting future work.

\section{Problem Settings}\label{section2}

This section states the problem setting of conversational contextual bandits. Suppose there is a finite set $\mathcal{A}$ of arms. Each arm $a \in \mathcal{A}$ represents an item to be recommended and is associated with a feature vector $\boldsymbol{x}_a \in \mathbb{R}^d$. Without loss of generality, the feature vectors are assumed to be normalized, i.e., $\norm{\boldsymbol{x}_a}_2=1$, ${\forall} a\in\mathcal{A}$. The agent interacts with a user in $T\in\mathbb{N}_{+}$ rounds, whose preference of items is represented by an \textit{unknown} vector $\boldsymbol{\theta}^* \in \mathbb{R}^d$, $\norm{\boldsymbol{\theta}^*}_2\leq 1$. 

At each round $t=1,2,...,T$, a subset of arms $\mathcal{A}_t \subseteq \mathcal{A}$ are available to the agent to choose from. Based on historical interactions, the agent selects an arm $a_t\in \mathcal{A}_t$, and receives a corresponding reward $r_{a_t,t}\in[0,1]$. The reward is assumed to be a linear function of the contextual vectors 
\begin{equation}
r_{a_t,t}=\boldsymbol{x}_{a_t}^{\top}\boldsymbol{\theta}^*+\epsilon_t\,,
\label{equation1}
\end{equation}
where $\epsilon_t$ is 1-sub-Gaussian random noise with zero mean.

Let $a_t^*\in{\arg\max}_{a\in\mathcal{A}_t}\boldsymbol{x}_{a}^{\top}\boldsymbol{\theta}^*$ denote an optimal arm with the largest expected reward at $t$. The learning objective is to minimize the cumulative regret
\begin{equation}
R(T)=\sum_{t=1}^{T}\boldsymbol{x}_{a_t^*}^{\top}\boldsymbol{\theta}^*-\sum_{t=1}^{T}\boldsymbol{x}_{a_t}^{\top}\boldsymbol{\theta}^*.
\label{equation2}
\end{equation}

The agent can also occasionally query the user's feedback on some conversational key-terms to help elicit user preferences. In particular, a ``key-term" is a keyword or topic related to a subset of arms. For example, the key-term \textit{sports} is related to the arms like basketball, football, swimming, etc.

Suppose there is a finite set $\cK$ of key-terms. The relationship between arms and key-terms is given by a weighted bipartite graph $(\mathcal{A},\mathcal{K},\boldsymbol{W})$, where $\boldsymbol{W}\triangleq\left[w_{a,k}\right]_{a\in\mathcal{A},k\in\mathcal{K}}$ represents the relationship between arms and key-terms, i.e., a key-term $k\in \cK$ is associated to an arm $a\in\cA$ with weight $w_{a,k}\ge 0$. We assume that each key-term $k$ has positive weights with some related arms (i.e., $\sum_{a\in\mathcal{A}}
w_{a,k}>0$, $\forall{k\in\mathcal{K}}$), and the weights associated with each arm sum up to 1, i.e., $\sum_{k\in\mathcal{K}} w_{a,k}=1$, $a\in \mathcal{A}$. The feature vector of a key-term $k$ is given by $\tilde{\boldsymbol{x}}_{k}=\sum_{a\in\mathcal{A}}\frac{w_{a,k}}{\sum_{a^{\prime}\in\mathcal{A}}w_{a^{\prime},k}}\boldsymbol{x}_a$. The key-term-level feedback on the key-term $k$ at $t$ is defined as
\begin{equation}
    \tilde{r}_{k,t}=\tilde{\boldsymbol{x}}_{k}^{\top}\boldsymbol{\theta}^*+\tilde{\epsilon}_t\,,
    \label{equation3}
\end{equation}
where $\tilde{\epsilon}_t$ is assumed to be 1-sub-Gaussian random noise. One thing to stress is that in the previous works \cite{zhang2020conversational,wu2021clustering,xie2021comparison,zhao2022knowledge}, the \textit{unknown} user preference vector $\boldsymbol{\theta}^*$ is essentially assumed to be the same at both the arm level and the key-term level.

To avoid affecting the user experience, the agent should not conduct conversations too frequently. Following \cite{zhang2020conversational}, we define a function $b: \mathbb{N}_+ \rightarrow \mathbb{R}_+$, where $b(t)$ is increasing in $t$, to control the conversation frequency of the agent. At each round $t$, if $b(t)-b(t-1)>0$, the agent is allowed to conduct $q(t)=\lfloor b(t)-b(t-1)\rfloor$ conversations by asking for user's feedback on $q(t)$ key-terms. Using this modeling arrangement, the agent will have $b(t)$ conversational interactions with the user up
to round $t$.

\section{Algorithms and Theoretical Analysis} \label{section3}

This section first introduces ConLinUCB, a framework for conversational bandits with better information incorporation, which is general for ``any" key-term selection strategies. Based on ConLinUCB, we further propose two bandit algorithms, ConLinUCB-BS and ConLinUCB-MCR, with explorative key-term selection strategies.

To simplify the exposition, we merge the ConLinUCB framework, ConLinUCB-BS and ConLinUCB-MCR in Algorithm \ref{algorithm1}. We also theoretically give regret bounds of our proposed algorithms.

\subsection{4.1 General ConLinUCB Algorithm Framework}
In conversational bandits, it is common that the \textit{unknown} preference vector $\boldsymbol{\theta}^*$ is essentially assumed to be the same at both arm level and key-term level  \cite{zhang2020conversational,xie2021comparison,wu2021clustering}. However, all existing works treat $\boldsymbol{\theta}^*$ differently at these two levels.
Specifically, they take two different steps
to estimate user preference vectors at the arm level and key-term
level, and use a discounting parameter $\lambda\in(0,1)$ to balance learning from these two levels' interactions. In this manner, the contributions of the arm-level and key-term-level information to the convergence of estimation are discounted by $\lambda$ and $1-\lambda$, respectively. Therefore, such discounting will cause waste of observations, indicating that information at these two levels can not be fully leveraged to accelerate the learning process.

To handle the above issues, we propose a general framework called ConLinUCB, for conversational contextual
bandits. In this new framework, in order to fully leverage interactive information from two levels, we simultaneously estimate the user preference vector by solving \textit{one
single optimization problem} that minimizes the mean squared error of both arm-level estimated rewards and key-term-level estimated feedback. Specifically, in ConLinUCB, at round $t$, the user preference vector is estimated by solving the following linear regression
   
        \begin{align}
            \boldsymbol{\theta}_t &=\mathop{\arg\min}\limits_{\boldsymbol{\theta}\in\mathbb{R}^d} \sum_{\tau=1}^{t-1} (\boldsymbol{x}^{\top}_{a_\tau}\boldsymbol{\theta}-r_{a_\tau,\tau})^2+\sum_{\tau=1}^{t} \sum_{k\in \mathcal{K}_\tau} ( \boldsymbol{\tilde{x}}^{\top}_{k}\boldsymbol{\theta} -\tilde{r}_{k,\tau})^2\nonumber\\\
            &\quad\quad+ \beta\norm{\boldsymbol{\theta}}_2^2\,,
        \label{equation4}
        \end{align}
where $\mathcal{K}_\tau$ denotes the set of key-terms asked at round $\tau$, and the coefficient $\beta>0$ controls regularization. The closed-form solution of this optimization problem is
    \begin{equation}
        \boldsymbol{\theta}_t=\boldsymbol{M}_t^{-1}\boldsymbol{b}_t\,,
    \label{equation5}
    \end{equation}
	where\begin{align}\begin{aligned}
        \boldsymbol{M}_t&=\sum_{\tau=1}^{t-1} \boldsymbol{x}_{a_\tau}\boldsymbol{x}^{\top}_{a_\tau}+\sum_{\tau=1}^{t} \sum_{k\in \mathcal{K}_\tau} \boldsymbol{\tilde{x}}_{k} \boldsymbol{\tilde{x}}^{\top}_{k}+\beta\boldsymbol{I}\,, \\
        		\boldsymbol{b}_t&=\sum_{\tau=1}^{t-1} \boldsymbol{x}_{a_\tau} r_{a_\tau,\tau}+\sum_{\tau=1}^{t} \sum_{k\in \mathcal{K}_\tau} \boldsymbol{\tilde{x}}_{k} \tilde{r}_{k,\tau}.
    \label{equation6}
    \end{aligned}\end{align}

\begin{algorithm}[t]
    \caption{General ConLinUCB framework}
    \label{algorithm1}
    \LinesNumbered
    \KwIn {graph$(\mathcal{A},\mathcal{K},\boldsymbol{W})$, conversation frequency function $b(t)$, key-term selection strategy $\boldsymbol{\pi}$.}
    \textbf{Initialization}: $\boldsymbol{M}_0=\beta\boldsymbol{I}$, $\boldsymbol{b}_0=\boldsymbol{0}$.\\
    \For{t = 1 to T}{   
        \eIf{$b(t)-b(t-1)>0$} {
            $q(t)=\lfloor b(t)-b(t-1)\rfloor$;\\
            \While{$q(t)>0$}{
            Select a key-term $k\in \mathcal{K}$ using the specified key-term selection strategy $\boldsymbol{\pi}$ (e.g., Eq. (\ref{BS}) for ConLinUCB-BS and Eq. (\ref{MCR}) for ConLinUCB-MCR), and query the user's preference over it;\\
            Receive the user's feedback $\tilde{r}_{k,t}$;\\
            $\boldsymbol{M}_t=\boldsymbol{M}_{t-1}+\tilde{\boldsymbol{x}}_{k}\tilde{\boldsymbol{x}}^{\top}_{k}$;\\
            $\boldsymbol{b}_t=\boldsymbol{b}_{t-1}+\tilde{\boldsymbol{x}}_{k}\tilde{r}_{k,t}$;\\
            $q(t)\mathrel{-}=1$;
            }
        } {
            $\boldsymbol{M}_t=\boldsymbol{M}_{t-1}$,
            $\boldsymbol{b}_t=\boldsymbol{b}_{t-1}$;
        }
$\boldsymbol{\theta}_t=\boldsymbol{M}_t^{-1}\boldsymbol{b}_t$;\\
Select $a_t = \mathop{\arg\max}\limits_{a \in \mathcal{A}_t} \boldsymbol{x}^{\top}_{a} \boldsymbol{\theta}_t +\alpha_t\norm{\boldsymbol{x}_{a}}_{\boldsymbol{M}_t^{-1}}$;\\
Ask the user's preference on arm $a_t$ and receive the reward $r_{a_t,t}$
;\\
$\boldsymbol{M}_t=\boldsymbol{M}_{t-1}+\boldsymbol{x}_{a_t}\boldsymbol{x}^{\top}_{a_t}$;\\
$\boldsymbol{b}_t=\boldsymbol{b}_{t-1}+\boldsymbol{x}_{a_t}r_{a_t,t}$;\\
    }
\end{algorithm}

To balance exploration and exploitation, ConLinUCB selects arms using the upper confidence bound (UCB) strategy
        \begin{equation}
                a_t = \mathop{\arg\max}\limits_{a \in \mathcal{A}_t} \underbrace{\boldsymbol{x}^{\top}_{a} \boldsymbol{\theta}_t}_{\hat{R}_{a,t}} +\underbrace{\alpha_t\norm{\boldsymbol{x}_{a}}_{\boldsymbol{M}_t^{-1}}}_{C_{a,t}}\,,
            \label{equation7}
        \end{equation}
        where $\norm{\boldsymbol{x}}_{\boldsymbol{M}}=\sqrt{\boldsymbol{x}^{\top}\boldsymbol{M}\boldsymbol{x}}$, $\hat{R}_{a,t}$ and $C_{a,t}$ denote the estimated reward and  confidence radius of arm $a$ at round $t$, and 
         \begin{equation}
        \alpha_t=\sqrt{2\log(\frac{1}{\delta})+d\log(1+\frac{t+b(t)}{\beta d})} +\sqrt{\beta}\,,
        \label{equation8}
        \end{equation}
        which comes from the following Lemma \ref{lemma2}.

The ConLinUCB algorithm framework is shown in Alg. \ref{algorithm1}. The key-term-level interactions take place in line 3-14. At round $t$, the agent first determines whether conversations are allowed using $b(t)$. When conducting conversations, the agent asks for the user's feedback on $q(t)$ key-terms and uses the feedback to update the parameters. Line 15-20 summarise the arm-level interactions. Based on historical interactions, the agent calculates the estimated $\boldsymbol{\theta}^*$, selects an arm with the largest UCB index, receives the corresponding reward, and updates the parameters accordingly. ConLinUCB only maintains one set of covariance matrix $\boldsymbol{M}_t$ and regressand vector $\boldsymbol{b}_t$, containing the feedback from both arm-level and key-term-level interactions. By doing so, ConLinUCB better leverages the feedback information than ConUCB. Note that ConLinUCB is a general framework with the specified key-term selection strategy $\boldsymbol{\pi}$ to be determined.

\subsection{4.2 ConLinUCB with key-terms from Barycentric Spanner (ConLinUCB-BS)}\label{section3.2}
Based on the ConLinUCB framework, we propose the ConLinUCB-BS algorithm with an explorative key-term selection strategy. Specifically, 
ConLinUCB-BS selects key-terms from the barycentric spanner $\mathcal{B}$ of the key-term set $\mathcal{K}$, which is an efficient exploration basis in online learning \cite{amballa2021computing}, to conduct explorative conversations. Below is the formal definition of the barycentric spanner for the key-term set $\mathcal{K}$.
\begin{definition}[Barycentric Spanner of $\mathcal{K}$]\label{def1}
A subset $\mathcal{B}=\{k_1,k_2,...,k_d\}
\subseteq\mathcal{K}$ is a barycentric spanner for $\mathcal{K}$ if for any $k\in\mathcal{K}$, there exists a set of coefficients $\boldsymbol{c}\in[-1,1]^d$, such that $\tilde{\boldsymbol{x}}_k=\sum_{i=1}^d\boldsymbol{c}_i\tilde{\boldsymbol{x}}_{k_i}$.
\end{definition}

We assume that the key-term set $\mathcal{K}$ is finite and $\{\tilde{\bx}_{k}\}_{k \in \cK}$ span $\mathbb{R}^d$, thus the existence of a barycentric spanner $\mathcal{B}$ of $\mathcal{K}$ is guaranteed \cite{awerbuch2008online}. 

Corresponding vectors in the barycentric spanner are linearly independent. By choosing key-terms from the barycentric spanner, we can quickly explore the \textit{unknown} user preference vector $\boldsymbol{\theta}^*$ in various directions. Based on this reasoning, whenever a conversation is allowed, ConLinUCB-BS selects a key-term
\begin{equation}
\label{BS}
    {k}\sim \text{unif}(\mathcal{B}),
\end{equation}
which means sampling a key-term $k$ uniformly at random from the barycentric spanner $\mathcal{B}$ of $\mathcal{K}$. ConLinUCB-BS is completed using the above strategy as $\boldsymbol{\pi}$ in the ConLinUCB framework (Alg. \ref{algorithm1}). As shown in the following Lemma \ref{lemma2} and Lemma \ref{lemma3}, in ConLinUCB-BS, the statistical estimation uncertainty shrinks quickly. Additionally, since the barycentric spanner $\mathcal{B}$ of the key-term set $\mathcal{K}$ can be precomputed offline, ConLinUCB-BS is computationally efficient, which is vital for real-time recommendations.

\subsection{4.3 ConLinUCB with key-terms having Max Confidence Radius (ConLinUCB-MCR)}\label{section3.3}
  We can further improve ConLinUCB-BS in the following aspects. First, ConLinUCB-BS does not apply in a more general setting where the key-term set $\mathcal{K}$ varies over time since it needs a precomputed barycentric spanner $\mathcal{B}$ of $\mathcal{K}$. Second, as the selection of key-terms is independent of past observations, ConLinUCB-BS does not fully leverage the historical information. For example, suppose the agent is already certain about whether the user favors \textit{sports} based on previous interactions. In that case, it does not need to ask for the user's feedback on the key-term \textit{sports} anymore. To address these issues, we propose the ConLinUCB-MCR algorithm that (i) is applicable when the key-term set $\cK$ varies with $t$ and (ii) can adaptively conduct explorative conversations based on historical interactions. 

In multi-armed bandits, confidence radius is used to capture whether an arm has been well explored in the interactive history, and it will shrink whenever the arm is selected. Motivated by this, if a key-term has a large confidence radius, it means the system has not sufficiently explored the user's preferences in its related items, indicating that this key-term is explorative. 
Based on this reasoning, ConLinUCB-MCR selects key-terms with maximal confidence radius to conduct explorative conversations apdaptively. Specifically, when a conversation is allowed at $t$, ConLinUCB-MCR chooses a key-term as follow
\begin{equation}
    \label{MCR}
    k\in\mathop{\arg\max}\limits_{k \in \mathcal{K}_t}\alpha_t\norm{\boldsymbol{\tilde{x}}_{k}}_{\boldsymbol{M}_t^{-1}}\,,
\end{equation}
where $\alpha_t$ is defined in Eq. (\ref{equation8}) and $\mathcal{K}_t\subseteq\mathcal{K}$ denotes the possibly time-varying  key-terms set available at round $t$. ConLinUCB-MCR is completed using the above strategy (Eq. (\ref{MCR})) as $\boldsymbol{\pi}$ in ConLinUCB (Alg. \ref{algorithm1}).

\subsection{4.4 Theoretical Analysis}
We give upper bounds of the regret for our algorithms. As a convention, the conversation frequency satisfies $b(t)\leq t$, so we assume $b(t)= b\cdot t$, $b\in(0,1)$. We leave the proofs of Lemma \ref{lemma2}-\ref{lemma3} and Theorem \ref{theorem1}-\ref{theorem4} to the Appendix due to the space limitation.

The following lemma shows a high probability upper bound of the difference between $\boldsymbol{\theta}_t$ and $\boldsymbol{\theta}^*$ in the direction of the action vector $\boldsymbol{x}_{a}$ for algorithms based on ConLinUCB.
\begin{figure*}[htbp]
	\centering
	\begin{subfigure}{0.49\linewidth}
		\centering
		\includegraphics[width=0.75\linewidth]{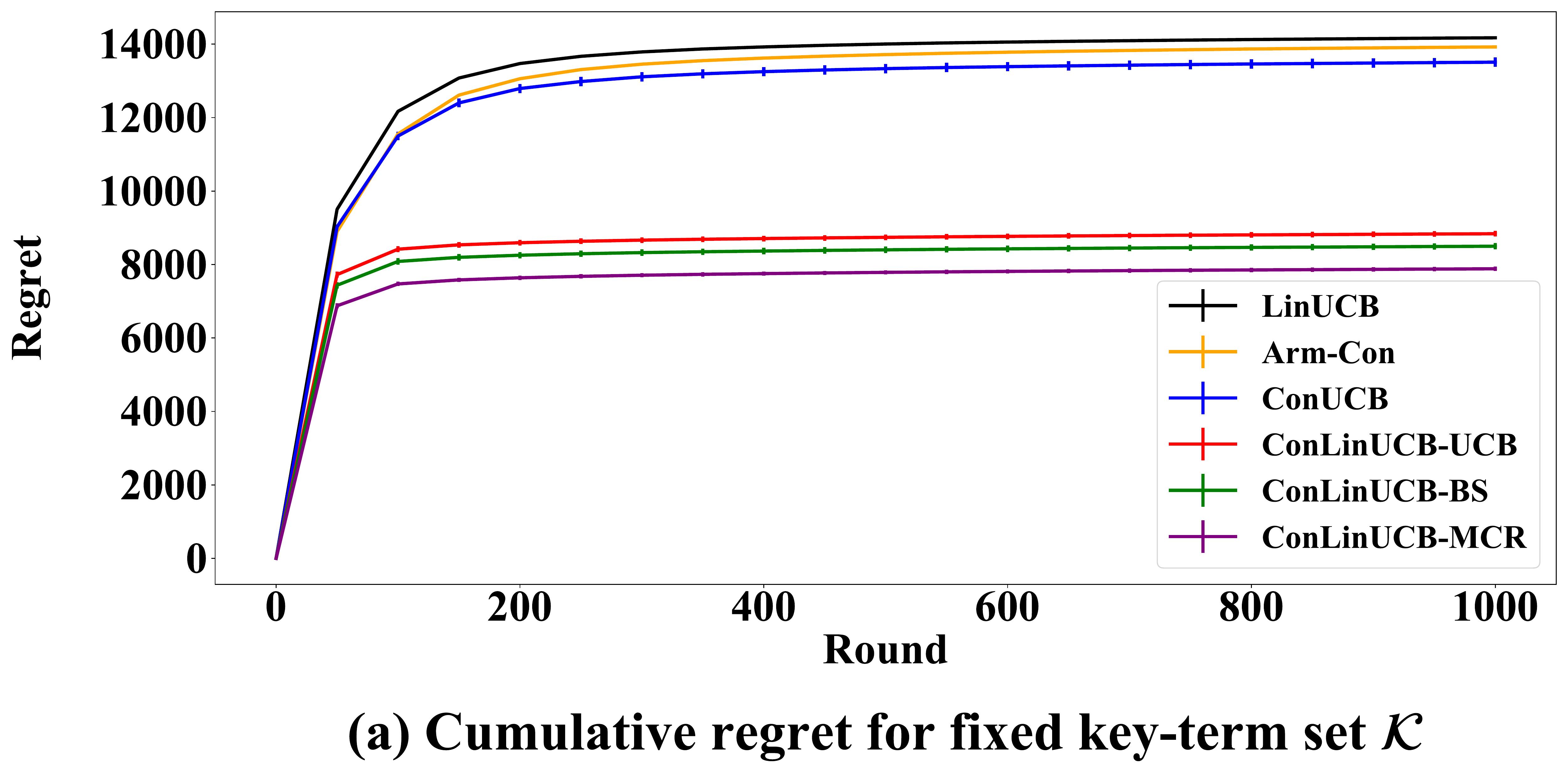}
	\end{subfigure}
	\centering
	\begin{subfigure}{0.49\linewidth}
		\centering
		\includegraphics[width=0.75\linewidth]{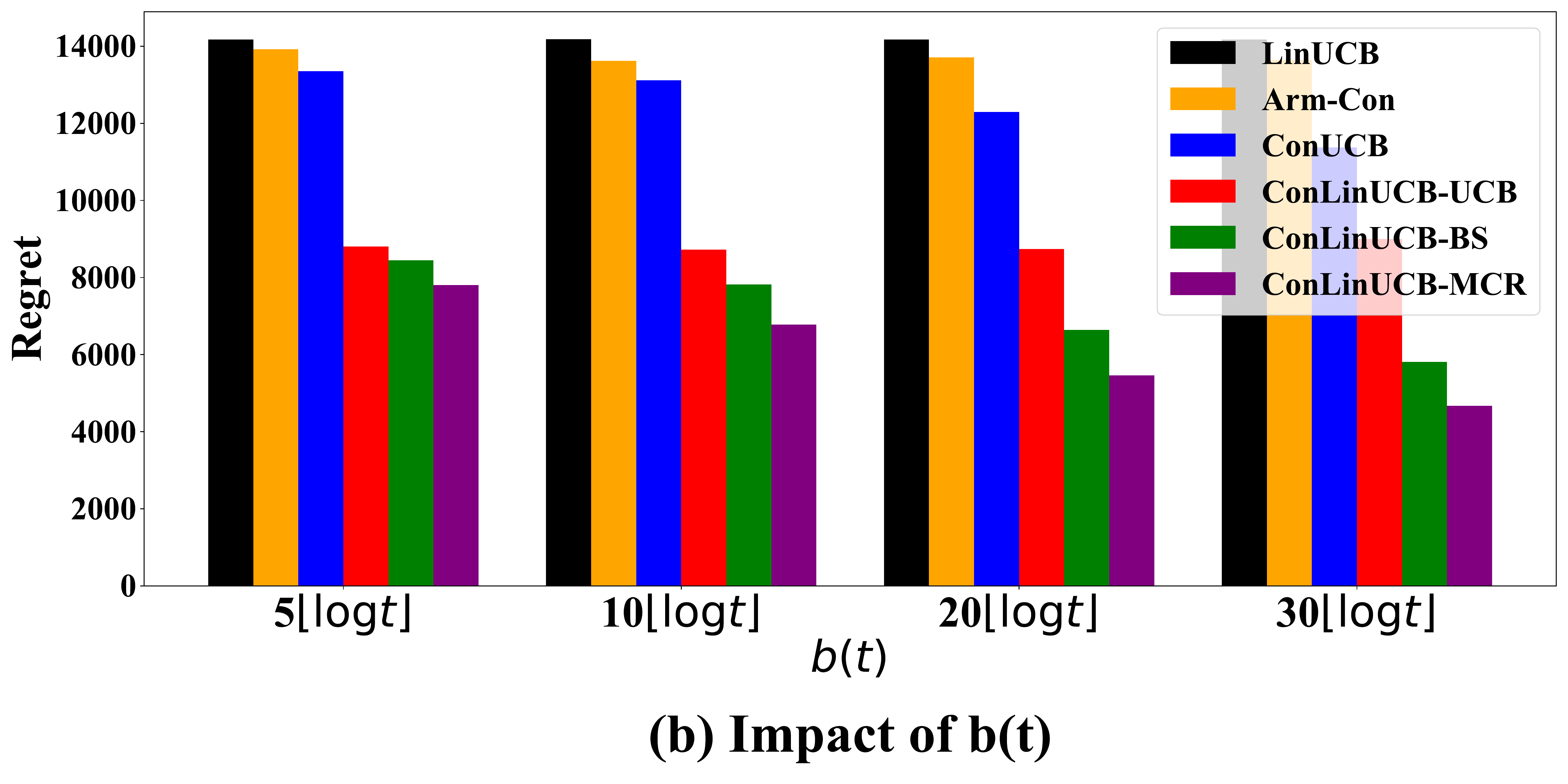}
	\end{subfigure}

	\centering
	\begin{subfigure}{0.49\linewidth}
		\centering
		\includegraphics[width=0.75\linewidth]{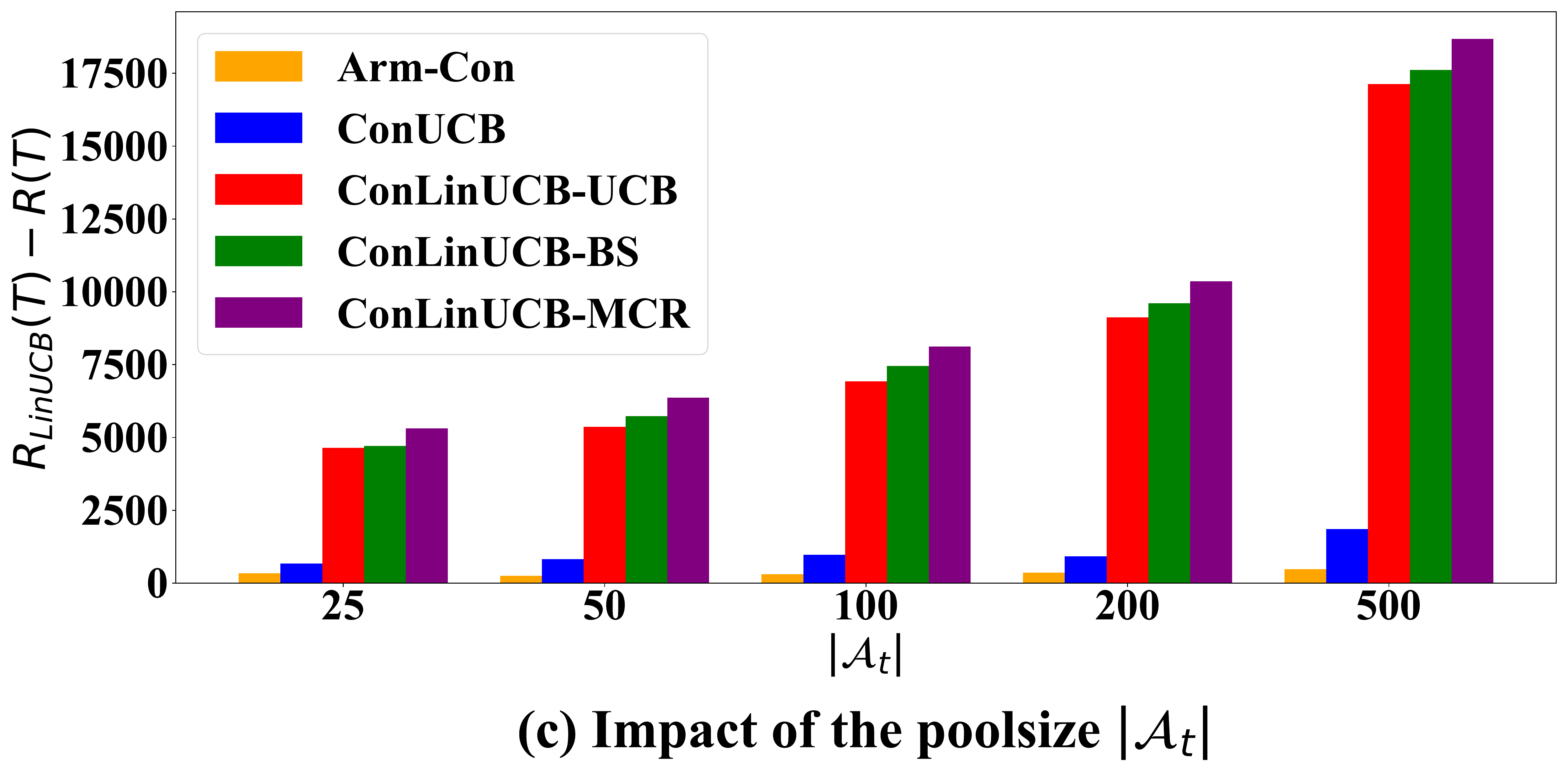}
	\end{subfigure}
		\begin{subfigure}{0.49\linewidth}
		\centering
		\includegraphics[width=0.75\linewidth]{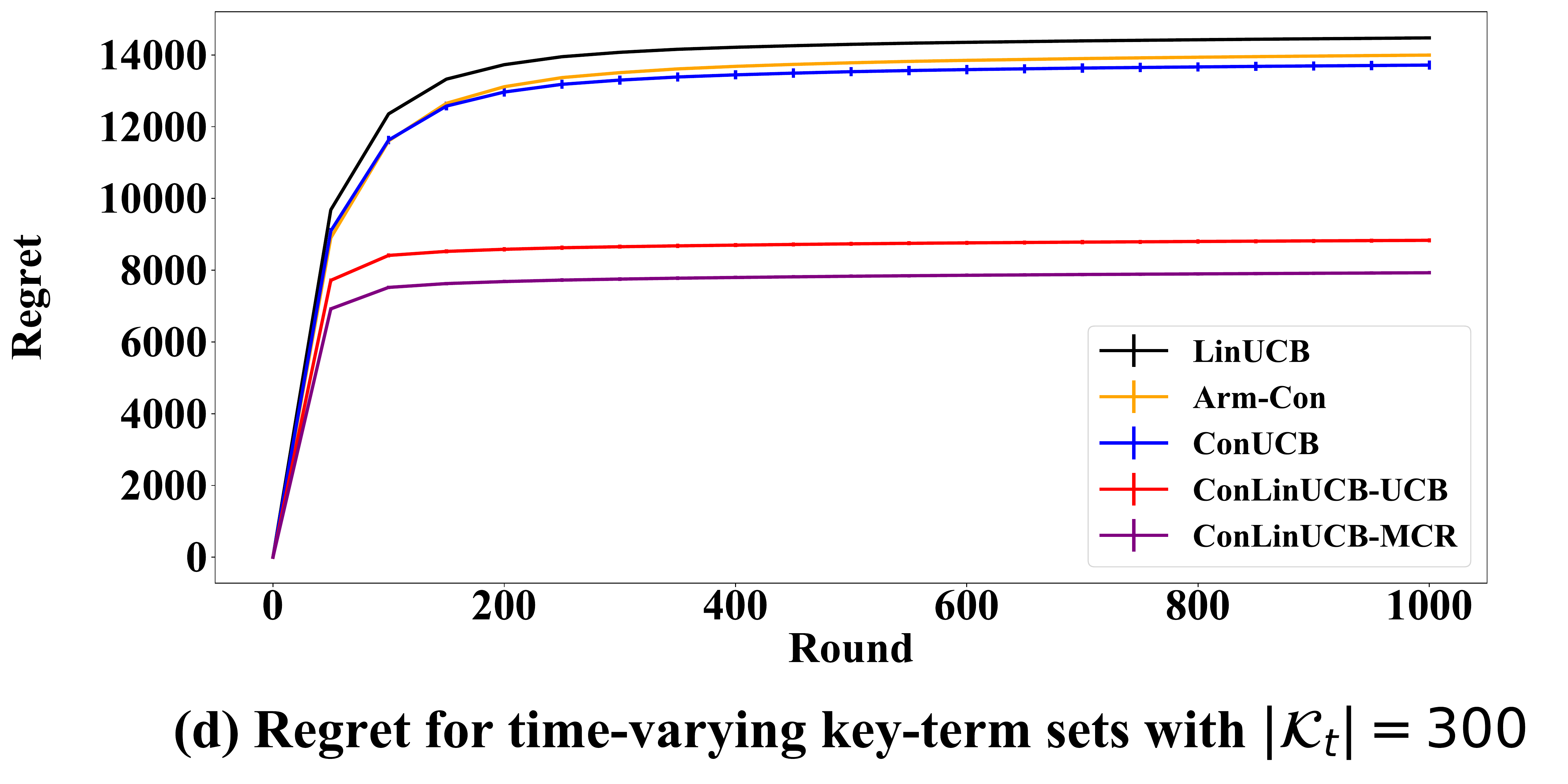}
	\end{subfigure}
	\caption{Experimental results on synthetic dataset} 
	\label{fig1}
\end{figure*}
\begin{lemma}\label{lemma2}
At ${\forall}t$, for any $a\in \mathcal{A}$, with probability at least $1-\delta$ for some $\delta \in (0,1)$
$$\left|\boldsymbol{x}_{a}^{\top}(\boldsymbol{\theta}_t-\boldsymbol{\theta}^*)\right|\leq\alpha_t\norm{\boldsymbol{x}_{a}}_{\boldsymbol{M}_t^{-1}}=C_{a,t},$$
where $\alpha_t=\sqrt{2\log(\frac{1}{\delta})+d\log(1+\frac{t+b(t)}{\beta d})} +\sqrt{\beta}$.
\end{lemma}

For a  barycentric spanner $\mathcal{B}$ of the key-term set $\mathcal{K}$, let
\begin{equation}
        \lambda_{\mathcal{B}}\coloneqq\lambda_{\min}(\boldsymbol{E}_{k\sim \text{unif}(\mathcal{B})}[\tilde{\boldsymbol{x}}_k\tilde{\boldsymbol{x}}_k^{\top}])>0\,,
        \label{equation11}
\end{equation}
where $\lambda_{\min}(\cdot)$ denotes the minimum eigenvalue of the augment. 
We can get the following Lemma that gives a high probability upper bound of $\norm{\boldsymbol{x}_{a}}_{\boldsymbol{M}_t^{-1}}$ for ConLinUCB-BS.
\begin{lemma}\label{lemma3}

For ConLinUCB-BS, ${\forall}a\in \mathcal{A}$, at ${\forall}t\geq t_0=\frac{256}{b\lambda_{\mathcal{B}}^2}\log(\frac{128d}{\lambda_{\mathcal{B}}^2\delta})$, with probability at least $1-\delta$ for $\delta \in (0,\frac{1}{8}]$
$$\norm{\boldsymbol{x}_{a}}_{\boldsymbol{M}_t^{-1}}\leq\sqrt{\frac{2}{\lambda_{\mathcal{B}} bt}}\,.$$
\end{lemma}

The following theorem gives a high probability regret upper bound of our ConLinUCB-BS.

\begin{theorem}\label{theorem1}
With probability at least $1-\delta$ for some $\delta \in (0,\frac{1}{4}]$, the regret $R(T)$ of ConLinUCB-BS satisfies
\begin{equation*}
    \begin{aligned}
    R(T)
    &\leq4\sqrt{\frac{2}{b\lambda_{\mathcal{B}}}}\sqrt{T}\Bigg(\sqrt{2\log(\frac{2}{\delta})+d\log(1+\frac{(1+b)T}{\beta d})}\\
    &\quad+\sqrt{\beta}\Bigg)+ \frac{256}{b\lambda_{\mathcal{B}}^2}\log(\frac{256d}{\lambda_{\mathcal{B}}^2\delta})+1\,.
    \end{aligned}
\end{equation*}
\end{theorem}

Recall that the regret upper bound of ConUCB \cite{zhang2020conversational} is
\begin{equation*}
    \begin{aligned}
    R(T) &\leq 2\sqrt{2Td\log(1+\frac{\lambda(T+1)}{(1-\lambda)d})}\Bigg(\sqrt{\frac{1-\lambda}{\lambda}}\\
        &\quad\ +\sqrt{\frac{1-\lambda}{\lambda\beta}}\sqrt{2\log(\frac{2}{\delta})+d\log(1+\frac{bT}{\beta d})}\\
        &\quad\ +\sqrt{2\log(\frac{2}{\delta})+d\log(1+\frac{\lambda T}{(1-\lambda) d})}\Bigg)\,,
    \end{aligned}
\end{equation*}
which is of $O(d\sqrt{T}\log T)$. The regret bound of ConLinUCB-BS given in Theorem \ref{theorem1} is of $O(d\sqrt{T\log T})$ (as  $\lambda_{\cB}$ is of order $O(\frac{1}{d})$), better than ConUCB by reducing a multiplicative $\sqrt{\log T}$ term.

Next, the following theorem gives a high-probability regret upper bound of ConLinUCB-MCR.
\begin{theorem}\label{theorem4}
With probability at least $1-\delta$ for some $\delta \in (0,1)$, the regret $R(T)$ of ConLinUCB-MCR satisfies
\begin{equation*}
    \begin{aligned}
    R(T)&\leq2\sqrt{2Td\log(1+\frac{T+1}{\beta d})}\\
    &\times\Bigg(\sqrt{\beta}+\sqrt{2\log(\frac{1}{\delta})+d\log(1+\frac{(b+1)T}{\beta d})}\Bigg)\,.
    \end{aligned}
\end{equation*}
\end{theorem}

Note that the regret upper bound of ConLinUCB-MCR is smaller than ConUCB by reducing some additive terms.

\section{Experiments on Synthetic Dataset} \label{section5}
In this section, we show the experimental results on synthetic data. To obtain the offline-precomputed barycentric spanner $\cB$, we use the method proposed in \cite{awerbuch2008online}.
\begin{figure*}[htbp]
	\centering
	\begin{subfigure}{0.49\linewidth}
		\centering
		\includegraphics[width=0.8\linewidth]{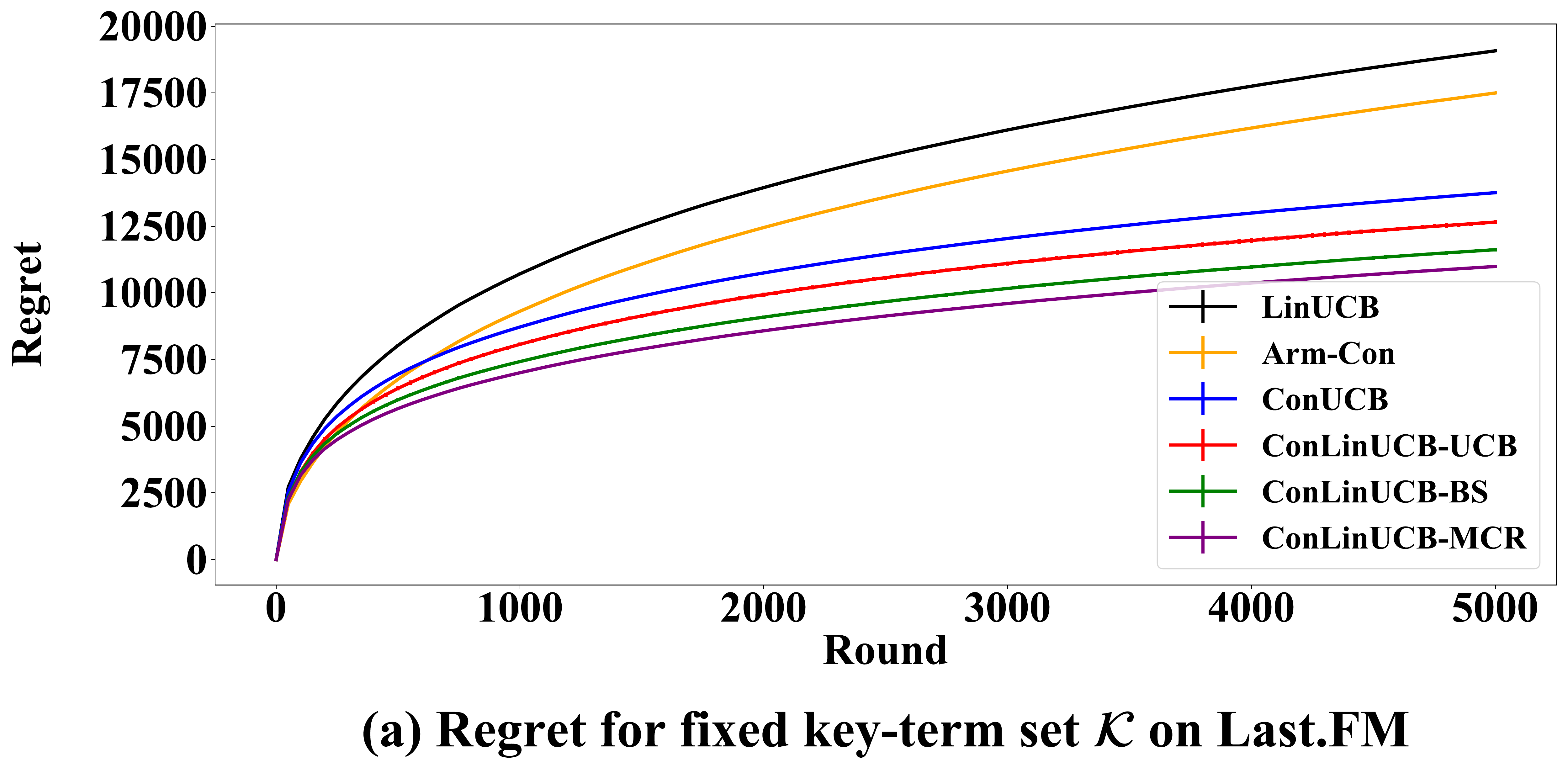}
	\end{subfigure}
	\centering
	\begin{subfigure}{0.49\linewidth}
		\centering
		\includegraphics[width=0.75\linewidth]{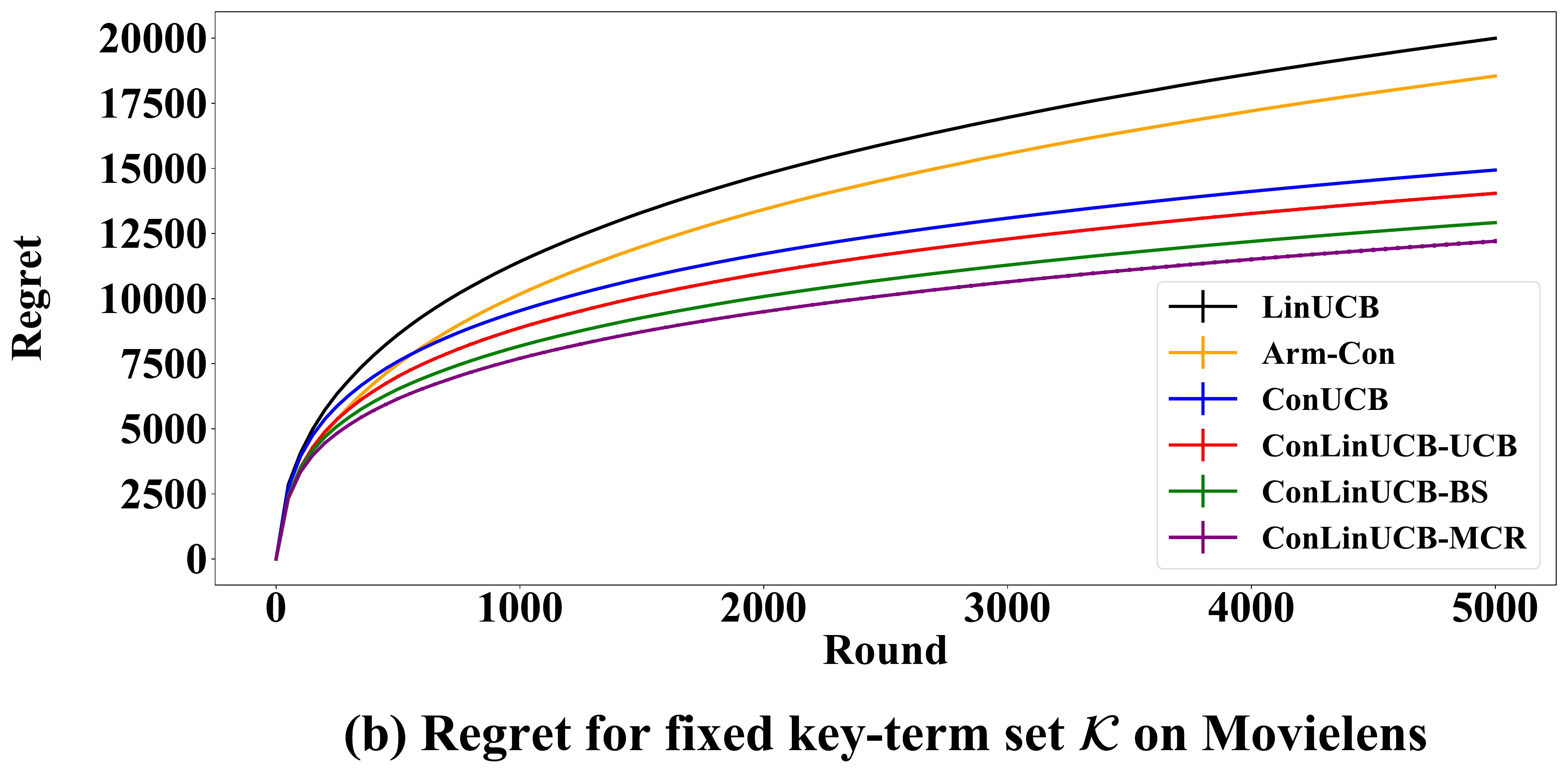}
	\end{subfigure}

	\centering
	\begin{subfigure}{0.49\linewidth}
		\centering
		\includegraphics[width=0.75\linewidth]{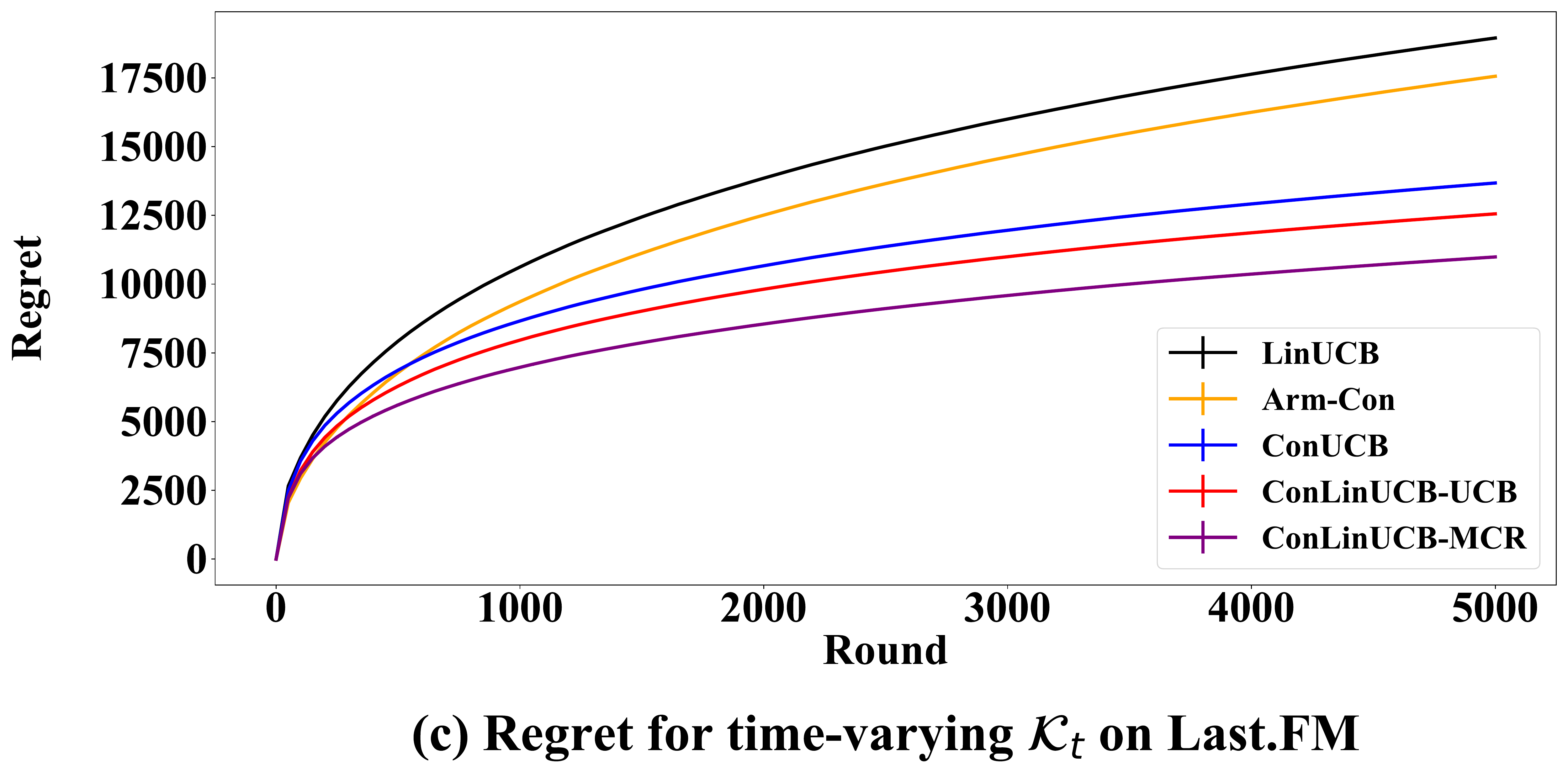}
	\end{subfigure}
		\begin{subfigure}{0.49\linewidth}
		\centering
		\includegraphics[width=0.75\linewidth]{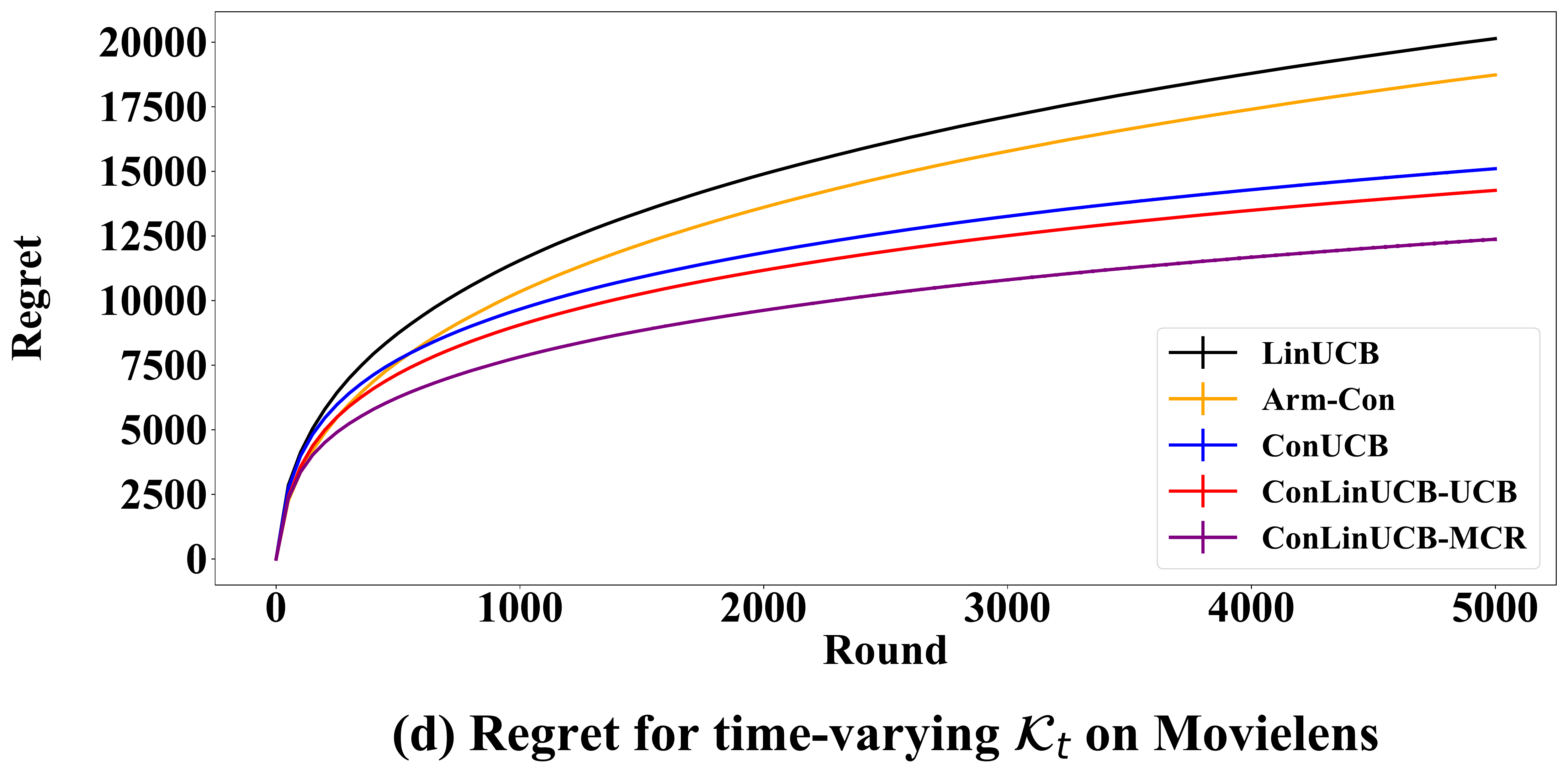}
	\end{subfigure}
	\caption{Experimental results on real-word datasets}
	\label{fig2}
\end{figure*}
\subsection{5.1 Experimental Settings}
\subsubsection{Generation of the synthetic dataset.}We create a set of arms $\mathcal{A}$ with $\left|\mathcal{A}\right|=5,000$ arms, and a set of key-terms $\mathcal{K}$ with $\left|\mathcal{K}\right|=500$. We set the dimension of the feature space to be $d=50$ and the number of users $N_u=200$.

For each user preference vector $\boldsymbol{\theta}_u^*$ and each arm feature vector $\boldsymbol{x}_a$, each entry is generated by independently drawing from the standard normal distribution $\mathcal{N}(0,1)$, and all these vectors are normalized such that $\norm{\boldsymbol{\theta}_u^*}_2=1$, $\norm{\boldsymbol{x}_a}_2=1$.
The weight matrix $\boldsymbol{W}\triangleq\left[w_{a,k}\right]$ is generated as follows: First, for each key-term $k$, we select an integer $n_k\in[1,10]$ uniformly at random, then randomly select a subset of $n_k$ arms $\mathcal{A}_k$ to be the related arms for key-term $k$; second, for each arm $a$, if it is related to a set of $n_a$ key-terms $\mathcal{K}_a$, we assign equal weights $w_{a,k}=\frac{1}{n_a}$, $\forall{k\in\mathcal{K}_a}$. Following \cite{zhang2020conversational}, the feature vector for each key-term $k$ is computed using
$\tilde{\boldsymbol{x}}_{k}=\sum_{a\in\mathcal{A}}\frac{w_{a,k}}{\sum_{a^{\prime}\in\mathcal{A}}w_{a^{\prime},k}}\boldsymbol{x}_a$. The arm-level rewards and key-term-level feedback are generated following Eq. (\ref{equation1}) and Eq. (\ref{equation3}). 
\subsubsection{Baselines.}\label{section5.1.2}We compare our algorithms with the following baselines:
\begin{itemize}
    \item LinUCB \cite{li2010contextual}: A state-of-the-art contextual linear bandit algorithm that selects arms only based on the arm-level feedback without using conversational feedback.
    \item Arm-Con \cite{christakopoulou2016towards}: A conversational bandit algorithm that conducts conversations on arms without considering key-terms, and uses LinUCB for arm selection. 
    \item ConUCB \cite{zhang2020conversational}: The core conversational bandit algorithm that selects a key-term to minimize some estimation error whenever a conversation is allowed.
    \item ConLinUCB-UCB: An algorithm using a LinUCB-alike method as the key-term selection strategy in our proposed ConLinUCB framework, i.e., choose key-term $    k\in\mathop{\arg\max}\limits_{k \in \mathcal{K}_t}\boldsymbol{\tilde{x}}_{k}^{\top}\boldsymbol{\theta}_t+\alpha_t\norm{\boldsymbol{\tilde{x}}_{k}}_{\boldsymbol{M}_t^{-1}}$ at round $t$.
\end{itemize}

\subsection{5.2 Evaluation Results}
This section first shows the results when the key-term set $\mathcal{K}$ is fixed. In this case, we evaluate the regret $R(T)$ for all algorithms, and we study the impact of the conversation frequency function $b(t)$ and the number of arms $\left|\mathcal{A}_t\right|$ available at each round $t$. When $\mathcal{K}$ varies with time, ConLinUCB-BS does not apply, and we compare the regret of other algorithms. Following \cite{zhang2020conversational}, we set $T=1,000$, $b(t)=5\lfloor\log (t+1)\rfloor$ and $\left|\mathcal{A}_t\right|=50$, unless otherwise stated. 

\begin{table*}
    \centering
    \resizebox{1.25\columnwidth}{!}{
	\begin{tabular}{ | c | c | c | c |}
		\hline
		Alogrithm 	& Total time 	& Total time for selecting arms & Total time for selecting key-terms \\ 
		\hline
		ConUCB 	& 11,297		& 5,217		 & 6,080  \\ 
		\hline 
		ConLinUCB-UCB & 5,738  		& 3,060		 & 2,678  \\ 
		\hline
		ConLinUCB-MCR & 4,821 		& 3,030		 & 1,791  \\
		\hline
		ConLinUCB-BS &  3,127		& 3,120		 & 6  \\ 
		\hline
	\end{tabular}
        }
            \caption{Total runninng time (in seconds) of algorithms on Movielens with $T=5,000$.}\label{table1}
\end{table*}
\subsubsection{Cumulative regret}
We run the experiments 10 times and calculate the average regret of all the users for each algorithm. We include $\pm std$ as the error bar, where $std$ stands for the \textit{standard deviation}. The results are given in Figure \ref{fig1} (a). First, all other algorithms outperform LinUCB, showing the advantage of conversations. Further, with our proposed ConLinUCB framework, even if we use ConLinUCB-UCB with a simple LinUCB-alike key-term selection strategy, the performance is already better than ConUCB (34.91\% improvement), showing more efficient information incorporation. With explorative conversations, ConLinUCB-BS and ConLinUCB-MCR achieve much lower regrets (37.00\% and 43.10\% improvement over ConUCB respectively), indicating better learning accuracy. ConLinUCB-MCR further leverages historical information to conduct explorative conversations adaptively, thus achieving the lowest regret.

\subsubsection{Impact of conversation frequency function $b(t)$} A larger $b(t)$ means the agent can conduct more conversations. We set $b(t)=f_q\cdot\lfloor\log t\rfloor$ and vary $f_q$ to change the conversation frequencies, i.e., $f_q\in\{5, 10, 20, 30\}$. The results are shown in Figure \ref{fig1} (b). With larger $b(t)$, our algorithms have less regret, showing the power of conversations. In all cases, ConLinUCB-BS and ConLinUCB-MCR have lower regrets than ConUCB, and ConLinUCB-MCR performs the best.

\subsubsection{Impact of $\left|\mathcal{A}_t\right|$} We vary $\left|\mathcal{A}_t\right|$ to be 25, 50, 100, 200, 500. To clearly show the advantage of our algorithms, we evaluate the difference in regrets between LinUCB and other algorithms, i.e., $R_{\text{LinUCB}}(T)-R(T)$, representing the improved accuracy of the conversational bandit algorithms as compared with LinUCB. Note that the larger $|\mathcal{A}_t|$ is, the harder it is for the algorithm to identify the best arm. Results in Figure \ref{fig1} (c) show that as $\left|\mathcal{A}_t\right|$ increases, the advantages of ConLinUCB-BS and ConLinUCB-MCR become more significant. Particularly, when $|\mathcal{A}_t|$=25, ConLinUCB-BS and ConLinUCB-MCR achieve 34.99\% and 40.21\% improvement over ConUCB respectively; when $|\mathcal{A}_t|$=500, ConLinUCB-BS and ConLinUCB-MCR achieve 50.36\% and 53.77\% improvement over ConUCB, respectively. In real applications, the size of arm set $\left|\mathcal{A}_t\right|$ is usually very large. Therefore, our proposed algorithms are expected to significantly outperform ConUCB in practice. 
\subsubsection{Cumulative regret for time-varying $\mathcal{K}$} 
This section studies the case when only a subset of key-terms $\mathcal{K}_t\subseteq\mathcal{K}$ are available to the agent at each round $t$, where ConLinUCB-BS is not applicable as mentioned before. The number of key-terms available at each time $t$ is set to be $\left|\mathcal{K}_t\right|=300$. At round $t$, 300 key-terms are chosen uniformly at random from $\mathcal{K}$ to form $\mathcal{K}_t$. We evaluate the regret of all algorithms except ConLinUCB-BS. The results are shown in Figure \ref{fig1} (d). We can observe that ConLinUCB-MCR outperforms all baselines and achieves 43.02\% improvement over ConUCB.

\section{Experiments on Real-world Datasets} \label{section6}
This section shows the experimental results on two real-world datasets, Last.FM and Movielens. The baselines, generations of arm-level rewards and key-term-level feedback, and the computation method of the barycentric spanner are the same as in the last section. Following the experiments on real data of \cite{zhang2020conversational}, we set $T=5,000$, $b(t)=5\lfloor\log (t+1)\rfloor$ and $\left|\mathcal{A}_t\right|=50$, unless otherwise stated.

\subsection{6.1 Experiment Settings}
\subsubsection{Last.FM and Movielens datasets \cite{Cantador:RecSys2011}}Last.FM is a dataset for music artist recommendations containing 186,479 interaction records between 1,892 users and 17,632 artists. Movielens is a dataset for movie recommendation containing 47,957 interaction records between 2,113 users and 10,197 movies.
\subsubsection{Generation of the data} The data is generated following \cite{10.5555/3367243.3367445,zhang2020conversational,wu2021clustering}. We treat each music artist and each movie as an arm. For both datasets, we extract $\left|\mathcal{A}\right|=2,000$ arms with the most assigned tags by users and $N_u=500$ users who have assigned the most tags. For each arm, we keep at most 20 tags that are related to the most arms, and consider them as the associated key-terms of the arm. All the kept key-terms associated with the arms form the key-term set $\mathcal{K}$. The number of key-terms for Last.FM is $\left|\mathcal{K}\right|=2,726$ and that for Movielens is $5,585$. The weights of all key-terms related to the same arm are set to be equal. Based on the interactive recordings, the user feedback is constructed as follows: if the user has assigned tags to the item, the feedback is 1, otherwise the feedback is 0. To generate the feature vectors of users and arms, following \cite{10.5555/3367243.3367445}, we construct a feedback matrix $\boldsymbol{F}\in \mathbb{R}^{N_u\times N}$ based on the above user feedback, and decompose it using the singular-value decomposition (SVD): $\boldsymbol{F}=\boldsymbol{\Theta}\boldsymbol{S}\boldsymbol{X}^{\top}$, where $\boldsymbol{\Theta}=(\boldsymbol{\theta_u^*})$, $u\in[N_u]$ and $\boldsymbol{X}=(\boldsymbol{x}_a)$, $a\in[N]$. We select $d=50$ dimensions with highest singular values in $\boldsymbol{S}$. Following \cite{zhang2020conversational}, feature vectors of key-terms are calculated using $\tilde{\boldsymbol{x}}_{k}=\sum_{a\in\mathcal{A}}\frac{w_{a,k}}{\sum_{a^{\prime}\in\mathcal{A}}w_{a^{\prime},k}}\boldsymbol{x}_a$.  The arm-level rewards and key-term-level feedback are then generated following Eq. (\ref{equation1}) and Eq. (\ref{equation3}).

\subsection{6.2 Evaluation Results}
This section first shows the results on both datasets in two cases: $\mathcal{K}$ is fixed and $\mathcal{K}$ is varying with time $t$. We also compare the running time of all algorithms on the Movielens dataset, since it has more key-terms than Last.FM.
\subsubsection{Cumulative regret}We run the experiments 10 times and calculate the average regret of all the users over $T=5,000$ rounds on the fixed generated datasets. The randomness of experiments comes from the randomly chosen $\mathcal{A}_t$ (also $\mathcal{K}_t$ in the varying key-term set case) and the randomness in the ConLinUCB-BS algorithm. We also include $\pm std$ as the error bar. For the time-varying key-term sets case, we set $\left|\mathcal{K}_t\right|=1,000$ and randomly select $\left|\mathcal{K}_t\right|$ key-terms from $\mathcal{K}$ to form $\mathcal{K}_t$ at round $t$. Results on Last.FM and Movielens for fixed key-term set are shown in Figure \ref{fig2} (a) and Figure \ref{fig2} (b). On both datasets, the regrets of ConLinUCB-BS and ConLinUCB-MCR are much smaller than ConUCB (13.28\% and 17.12\% improvement on Last.FM, 13.08\% and 16.93\% improvement on Movielens, respectively) and even the simple ConLinUCB-UCB based on our ConLinUCB framework outperforms ConUCB. Results on Last.FM and Movielens for varying key-term sets are given in Figure \ref{fig2} (c) and Figure \ref{fig2} (d). ConLinUCB-MCR performs much better than ConUCB on both datasets (19.66\% and 17.85\% improvement on Last.FM and Movielens respectively).

\subsubsection{Running time} We evaluate the running time of all the conversational bandit algorithms on the representative Movielens dataset to compare their computational efficiency. For clarity, we report the total running time for selecting arms and key-terms. We set $T=5,000$ and the results are summarized in Table \ref{table1}. It is clear that our algorithms cost much less time in both key-term selection and arm selection than ConUCB. Specifically, the improvements of total running time over ConUCB are 72.32\% for ConLinUCB-BS and 57.32\% for ConLinUCB-MCR. The main reason is that our algorithms estimate the \textit{unknown} user preference vector in one single step, whereas ConUCB does it in two separate steps as mentioned before. For ConLinUCB-BS, the time costed in the key-term selection is almost negligible, since it just randomly chooses a key-term from the precomputed barycentric spanner whenever a conversation is allowed.

\section{Conclusion}\label{section8}
In this paper, we introduce ConLinUCB, a general framework for conversational bandits with efficient information incorporation. Based on this framework, we propose ConLinUCB-BS and ConLinUCB-MCR,
with explorative key-term selection strategies that can quickly elicit the user's potential interests. We prove tight regret bounds of our algorithms. Particularly, ConLinUCB-BS achieves a bound
of $O(d\sqrt{T\log T})$, much better than $O(d\sqrt{T}\log T)$ of the classic ConUCB. In the empirical evaluations, our algorithms dramatically outperform the classic ConUCB. For future work, it would be interesting to consider the settings with knowledge graphs \cite{zhao2022knowledge}, hierarchy item trees \cite{song2022show}, relative feedback \cite{xie2021comparison} or different feedback selection strategies \cite{letard2020partial,letard2022mabs}, and use our framework and principles to improve the performance of existing algorithms.

\clearpage
\input{acknowledgement.tex}
\bibliography{aaai23}
\newpage
\appendix
\onecolumn
\section{Proof of Lemma \ref{lemma2}}
\begin{proof}
According to the closed-form solution of $\boldsymbol{\theta}_t$ in Eq. (\ref{equation5}) (\ref{equation6}), we can calculate the estimation error as follows
\begin{equation*}
\begin{aligned}
        \boldsymbol{\theta}_t-\boldsymbol{\theta}_*
        &=\boldsymbol{M}_t^{-1}\boldsymbol{b}_t-\boldsymbol{\theta}_*\\
        &=\Bigg(\sum_{\tau=1}^{t-1} \boldsymbol{x}_{a_\tau}\boldsymbol{x}^{\top}_{a_\tau}+\sum_{\tau=1}^{t} \sum_{k\in \mathcal{K}_\tau} \boldsymbol{\tilde{x}}_{k} \boldsymbol{\tilde{x}}^{\top}_{k}+\beta\boldsymbol{I}\Bigg)^{-1}\Bigg(\sum_{\tau=1}^{t-1} \boldsymbol{x}_{a_\tau} r_{a_\tau,\tau}+\sum_{\tau=1}^{t} \sum_{k\in \mathcal{K}_\tau} \boldsymbol{\tilde{x}}_{k} \tilde{r}_{k,\tau}\Bigg)-\boldsymbol{\theta}_*\\
        &=\Bigg(\sum_{\tau=1}^{t-1} \boldsymbol{x}_{a_\tau}\boldsymbol{x}^{\top}_{a_\tau}+\sum_{\tau=1}^{t} \sum_{k\in \mathcal{K}_\tau} \boldsymbol{\tilde{x}}_{k} \boldsymbol{\tilde{x}}^{\top}_{k}+\beta\boldsymbol{I}\Bigg)^{-1}\Bigg(\sum_{\tau=1}^{t-1} \boldsymbol{x}_{a_\tau} \bigg(\boldsymbol{x}^{\top}_{a_\tau}\boldsymbol{\theta}_*+\epsilon_\tau\bigg)
        +\sum_{\tau=1}^{t} \sum_{k\in \mathcal{K}_\tau} \boldsymbol{\tilde{x}}_{k} \bigg(\boldsymbol{\tilde{x}}_{k}^{\top}\boldsymbol{\theta}_*+\tilde{\epsilon}_{\tau}\bigg)\Bigg)-\boldsymbol{\theta}_*\\
        &=\Bigg(\sum_{\tau=1}^{t-1} \boldsymbol{x}_{a_\tau}\boldsymbol{x}^{\top}_{a_\tau}+\sum_{\tau=1}^{t} \sum_{k\in \mathcal{K}_\tau} \boldsymbol{\tilde{x}}_{k} \boldsymbol{\tilde{x}}^{\top}_{k}+\beta\boldsymbol{I}\Bigg)^{-1}\Bigg(\sum_{\tau=1}^{t-1} \boldsymbol{x}_{a_\tau}\boldsymbol{x}^{\top}_{a_\tau}+\sum_{\tau=1}^{t} \sum_{k\in \mathcal{K}_\tau} \boldsymbol{\tilde{x}}_{k} \boldsymbol{\tilde{x}}^{\top}_{k}+\beta\boldsymbol{I}-\beta\boldsymbol{I}\Bigg)\boldsymbol{\theta}_*-\boldsymbol{\theta}_*\\
        &\quad\quad+\boldsymbol{M}_t^{-1}(\sum_{\tau=1}^{t-1}\boldsymbol{x}_{a_\tau}\epsilon_\tau+\sum_{\tau=1}^{t} \sum_{k\in \mathcal{K}_\tau} \boldsymbol{\tilde{x}}_{k} \tilde{\epsilon}_{\tau})\\
        &=-\beta\boldsymbol{M}_t^{-1}\boldsymbol{\theta}_*+\boldsymbol{M}_t^{-1}(\sum_{\tau=1}^{t-1}\boldsymbol{x}_{a_\tau}\epsilon_\tau+\sum_{\tau=1}^{t} \sum_{k\in \mathcal{K}_\tau} \boldsymbol{\tilde{x}}_{k} \tilde{\epsilon}_{\tau})\,.
\end{aligned}
\end{equation*}	

We can then bound the projection of the estimation error onto the direction of the action vector $\boldsymbol{x}_{a}$:
\begin{align}
        \left|\boldsymbol{x}_{a}^{\top}(\boldsymbol{\theta}_t-\boldsymbol{\theta}_*)\right|&\leq\beta\left|\boldsymbol{x}_{a}^{\top}\boldsymbol{M}_t^{-1}\boldsymbol{\theta}_*\right| +\left|\boldsymbol{x}_{a}^{\top}\boldsymbol{M}_t^{-1}(\sum_{\tau=1}^{t-1}\boldsymbol{x}_{a_\tau}\epsilon_\tau+\sum_{\tau=1}^{t} \sum_{k\in \mathcal{K}_\tau} \boldsymbol{\tilde{x}}_{k} \tilde{\epsilon}_{\tau})\right|\nonumber
        \\
        &\leq\beta\norm{\boldsymbol{x}_{a}^{\top}\boldsymbol{M}_t^{-\frac{1}{2}}}_2\norm{\boldsymbol{M}_t^{-\frac{1}{2}}\boldsymbol{\theta}_*}_2 +\norm{\boldsymbol{x}_{a}^{\top}\boldsymbol{M}_t^{-\frac{1}{2}}}_2 \times\norm{\boldsymbol{M}_t^{-\frac{1}{2}}(\sum_{\tau=1}^{t-1}\boldsymbol{x}_{a_\tau}\epsilon_\tau+\sum_{\tau=1}^{t} \sum_{k\in \mathcal{K}_\tau} \boldsymbol{\tilde{x}}_{k} \tilde{\epsilon}_{\tau})}_2\label{cauchy1}\\
        &\leq\beta\norm{\boldsymbol{x}_{a}}_{\boldsymbol{M}_t^{-1}}\norm{\boldsymbol{M}_t^{-\frac{1}{2}}}_2\norm{\boldsymbol{\theta}_*}_2 +\norm{\boldsymbol{x}_{a}}_{\boldsymbol{M}_t^{-1}}\norm{\sum_{\tau=1}^{t-1}\boldsymbol{x}_{a_\tau,\tau}\epsilon_\tau+\sum_{\tau=1}^{t} \sum_{k\in \mathcal{K}_\tau} \boldsymbol{\tilde{x}}_{k} \tilde{\epsilon}_{\tau}}_{\boldsymbol{M}_t^{-1}}\label{operator norm1}\\
        &\leq\norm{\boldsymbol{x}_{a}}_{\boldsymbol{M}_t^{-1}}\Bigg(\sqrt{\beta}\norm{\boldsymbol{\theta}_*}_2+\norm{\sum_{\tau=1}^{t-1}\boldsymbol{x}_{a_\tau}\epsilon_\tau+\sum_{\tau=1}^{t} \sum_{k\in \mathcal{K}_\tau} \boldsymbol{\tilde{x}}_{k} \tilde{\epsilon}_{\tau}}_{\boldsymbol{M}_t^{-1}}\Bigg)\label{minimal eigen}\,,
        \end{align}
where Eq. (\ref{cauchy1}) is by the Cauchy–Schwarz inequality, Eq. (\ref{operator norm1}) is by the inequality of the matrix operator norm, and Eq. (\ref{minimal eigen}) is because $\lambda_{min}(\boldsymbol{M}_t)\geq\beta,\,\, \norm{\boldsymbol{M}_t^{-\frac{1}{2}}}_2=\sqrt{\lambda_{max}(\boldsymbol{M}_t^{-1})}=\sqrt{\frac{1}{\lambda_{min}(\boldsymbol{M}_t)}}\leq\sqrt{\frac{1}{\beta}}$.

Theorem 1 in \cite{abbasi2011improved} suggests that with probability at least $1-\delta$
\begin{equation}
      \norm{\sum_{\tau=1}^{t-1}\boldsymbol{x}_{a_\tau}\epsilon_\tau+\sum_{\tau=1}^{t} \sum_{k\in \mathcal{K}_\tau} \boldsymbol{\tilde{x}}_{k} \tilde{\epsilon}_{k}}_{\boldsymbol{M}_t^{-1}}
    \leq \sqrt{2\log\bigg(\frac{det(\boldsymbol{M}_t)^{\frac{1}{2}}det(\beta\boldsymbol{I})^{\frac{1}{2}}}{\delta}\bigg)}
      \label{equation need det}\,,
\end{equation}
where $det(\cdot)$ denotes the determinate of the argument.

We have
\begin{align}
        det(\boldsymbol{M}_t)= \prod_{i=1}^{d}\lambda_i
        &\leq\big(\frac{\sum_{i=1}^d \lambda_i}{d}\big)^d\label{mean inequality}\\
        &=\big(\frac{trace(\boldsymbol{M}_t)}{d}\big)^d \label{trace equality}\\
        &=\Bigg(\frac{trace(\sum_{\tau=1}^{t-1} \boldsymbol{x}_{a_\tau}\boldsymbol{x}^{\top}_{a_\tau}+\sum_{\tau=1}^{t} \sum_{k\in \mathcal{K}_\tau} \boldsymbol{\tilde{x}}_{k} \boldsymbol{\tilde{x}}^{\top}_{k}+\beta\boldsymbol{I})}{d}\Bigg)^d\nonumber\leq \big(\frac{t+b(t)+\beta d}{d}\big)^d\nonumber\,,
\end{align}
where $\lambda_i,i= 1,2,\ldots,d$ denotes the eigenvalues of the matrix $\boldsymbol{M}_t$, $trace(\boldsymbol{M}_t)$ denotes the trace of $\boldsymbol{M}_t$, Eq. (\ref{mean inequality}) follows by the inequality of arithmetic and geometric means, Eq. (\ref{trace equality}) follows since the trace of a matrix is equal to the sum of its eigenvalues.

Plugging the above inequality and $det(\beta\bI)=\beta^d$ into Eq. (\ref{equation need det}), we can get
\begin{equation}
        \norm{\sum_{\tau=1}^{t-1}\boldsymbol{x}_{a_\tau}\epsilon_\tau+\sum_{\tau=1}^{t} \sum_{k\in \mathcal{K}_\tau} \boldsymbol{\tilde{x}}_{k} \tilde{\epsilon}_{k}}_{\boldsymbol{M}_t^{-1}}\leq\sqrt{2\log(\frac{1}{\delta})+d\log(1+\frac{b(t)+t}{\beta d})}\label{bound on a term}\,.
\end{equation}
The result then follows by plugging Eq. (\ref{bound on a term}) into Eq. (\ref{minimal eigen}), and the fact that $\norm{\boldsymbol{\theta}^*}_2\leq 1$.
\end{proof}

\section{Proof of Lemma \ref{lemma3}}
\begin{proof}
Recall that in ConLinUCB-BS, the key-terms are uniformly sampled from the pre-computed barycentric spanner $\cB$, i.e., ${k}\sim \text{unif}(\mathcal{B})$. Therefore we have
\begin{equation}
    \lambda_{\mathcal{B}}\coloneqq\lambda_{\min}(\boldsymbol{E}_{k\sim \text{unif}(\mathcal{B})}[\tilde{\boldsymbol{x}}_k\tilde{\boldsymbol{x}}_k^{\top}])>0\,.
\end{equation}

Using Eq. (\ref{equation11}) in the Lemma 7 in \cite{li2018online}, and the fact that $b_t=b\cdot t$, then with probability at least $1-\delta$ for $\delta\in(0,\frac{1}{8}]$, we have
\begin{equation}
    \label{equation12}
        \lambda_{\min}(\sum_{\tau=1}^t\sum_{k \in \mathcal{K}_{\tau}}\boldsymbol{\tilde{x}}_{k}\boldsymbol{\tilde{x}}_{k}^{\top})\geq \frac{\lambda_{\mathcal{B}} bt}{2}\,,
\end{equation}
for all $t\geq t_0=\frac{256}{b\lambda_{\mathcal{B}}^2}\log(\frac{128d}{\lambda_{\mathcal{B}}^2\delta})$.

Then, by Courant–Fischer theorem \cite{ikebe1987monotonicity}, the fact that $\norm{\boldsymbol{x}_{a}}_2=1$, together with Eq. (\ref{equation12}), we have that for any $t\geq t_0$,with probability at least $1-\delta$ for $\delta\in(0,\frac{1}{8}]$,
\begin{equation*}
    \begin{aligned}
    \norm{\boldsymbol{x}_{a}}_{\boldsymbol{M}_t^{-1}}
    &=\sqrt{\boldsymbol{x}_{a}^{\top}\boldsymbol{M}_t^{-1}\boldsymbol{x}_{a}}\\
    &\leq \max_{\boldsymbol{x}\in\mathbb{R}^d\atop\norm{\boldsymbol{x}}_2=1}\sqrt{\boldsymbol{x}^{\top}\boldsymbol{M}_t^{-1}\boldsymbol{x}}\\
    &=\sqrt{\lambda_{\max}(\boldsymbol{M}_t^{-1})}\\
    &=\sqrt{\lambda_{\max}\bigg((\sum_{\tau=1}^{t-1} \boldsymbol{x}_{a_\tau}\boldsymbol{x}^{\top}_{a_\tau}+\sum_{\tau=1}^{t} \sum_{k\in \mathcal{K}_\tau} \boldsymbol{\tilde{x}}_{k} \boldsymbol{\tilde{x}}^{\top}_{k}+\beta\boldsymbol{I})^{-1}\bigg)}\\
    &=\sqrt{\frac{1}{\lambda_{\min}\bigg(\sum_{\tau=1}^{t-1} \boldsymbol{x}_{a_\tau}\boldsymbol{x}^{\top}_{a_\tau}+\sum_{\tau=1}^{t} \sum_{k\in \mathcal{K}_\tau} \boldsymbol{\tilde{x}}_{k} \boldsymbol{\tilde{x}}^{\top}_{k}+\beta\boldsymbol{I}\bigg)}}\\
    &\leq\sqrt{\frac{1}{\lambda_{\min}(\sum_{\tau=1}^t\sum_{k \in \mathcal{K}_{\tau}}\boldsymbol{\tilde{x}}_{k}\boldsymbol{\tilde{x}}_{k}^{\top})}}\\
    &\leq \sqrt{\frac{2}{\lambda_{\mathcal{B}} bt}}\,.
    \end{aligned}
\end{equation*}
\end{proof}

\section{Proof of Theorem \ref{theorem1}}
\begin{proof}
We denote the instantaneous regret at round $t$ as $R_t$. With the definition of the cumulative regret given in Eq. (\ref{equation2}), the arm selection strategy shown in Eq. (\ref{equation7}) and Lemma \ref{lemma2}, we can bound the regret $R_t$ at each round $t=1,2,3,4,...,T$ as follows
\begin{equation}
        \begin{aligned}
        R_t &= \boldsymbol{x}_{a_t^*}^{\top}\boldsymbol{\theta}^*- \boldsymbol{x}_{a_t}^{\top}\boldsymbol{\theta}^*\\
        &=\boldsymbol{x}_{a_t^*}^{\top}(\boldsymbol{\theta}^*-\boldsymbol{\theta}_t)+(\boldsymbol{\theta}_t^{\top}\boldsymbol{x}_{a_t^*}+C_{a_t^*,t})-(\boldsymbol{\theta}_t^{\top}\boldsymbol{x}_{a_t}+C_{a_t,t})\\
        & \quad\ +\boldsymbol{x}_{a_t}^{\top}(\boldsymbol{\theta}_t-\boldsymbol{\theta}^*)+C_{a_t,t}-C_{a_t^*,t}\\
        &\leq 2C_{a_t,t}\,.
        \end{aligned} 
        \label{equation13}
\end{equation}

With Lemma \ref{lemma3}, together with the assumption that $r_t \leq 1$ for any $t$, with probability at least $1-\delta$ for some $\delta \in (0,\frac{1}{4}]$, we can get
\begin{align}
    R(T)&=R(\lceil t_0 \rceil)+\sum_{t=\lceil t_0 \rceil +1}^T R_t\nonumber\\
    &\leq t_0+1+2\sum_{t=\lceil t_0 \rceil}^{T} C_{a_t,t}\nonumber\\
    &\leq t_0+1+2\alpha_t\sum_{t=\lceil t_0 \rceil} ^{T}\norm{\boldsymbol{x}_{a_t}}_{\boldsymbol{M}_t^{-1}}\nonumber\\
    &\leq t_0+1+2\alpha_T\sum_{t=\lceil t_0 \rceil} ^{T}\norm{\boldsymbol{x}_{a_t}}_{\boldsymbol{M}_t^{-1}}\label{nondecreasing}\\
    &\leq t_0+1+2\alpha_T \sum_{t=\lceil t_0 \rceil} ^{T} \sqrt{\frac{2}{\lambda_{\mathcal{B}} bt}}\nonumber\\
    &\leq t_0+1+2\alpha_T \sqrt{\frac{2}{\lambda_{\mathcal{B}}b}}\int_{t_0}^{T} \sqrt{\frac{1}{t}} dt\nonumber\\
    &=\leq t_0+1+4\alpha_T \sqrt{\frac{2}{\lambda_{\mathcal{B}}b}}(\sqrt{T}-\sqrt{t_0})\nonumber\\
    &\leq t_0+1+4\alpha_T \sqrt{\frac{2}{\lambda_{\mathcal{B}}b}}\sqrt{T}\nonumber\,,
\end{align}
where Eq. (\ref{nondecreasing}) follows since $\alpha_t$ is non-decreasing in $t$.

The result follows by plugging in the definition of $t_0$ and $\alpha_T$.
\end{proof}

\section{Proof of Theorem \ref{theorem4}}
\begin{proof}
We first prove the following result:\\
For any two positive definite matrices $\boldsymbol{A},\boldsymbol{B}\in\mathbb{R}^{d\times d}$, and any vector $\boldsymbol{x}\in\mathbb{R}^d$, we have:
\begin{equation}
    \norm{\boldsymbol{x}}_{(\boldsymbol{A}+\boldsymbol{B})^{-1}}^2\leq \norm{\boldsymbol{x}}_{\boldsymbol{A}^{-1}}^2\,.
    \label{psd inequality}
\end{equation}

This result can be proved by the following arguments:
\begin{align}
    \norm{\boldsymbol{x}}_{(\boldsymbol{A}+\boldsymbol{B})^{-1}}^2
    &=\boldsymbol{x}^{\top}(\boldsymbol{A}+\boldsymbol{B})^{-1}\boldsymbol{x}\nonumber\\
    &=\boldsymbol{x}^{\top}\big(\boldsymbol{A}^{-1}-\boldsymbol{A}^{-1}(\boldsymbol{B}^{-1}+\boldsymbol{A}^{-1})^{-1}\boldsymbol{A}^{-1}\big)\boldsymbol{x}\label{woodbury matrix identity}\\
    &=\boldsymbol{x}^{\top}\boldsymbol{A}^{-1}\boldsymbol{x}-(\boldsymbol{A}^{-1}\boldsymbol{x})^{\top}(\boldsymbol{B}^{-1}+\boldsymbol{A}^{-1})^{-1}(\boldsymbol{A}^{-1}\boldsymbol{x})\nonumber\\
    &\leq \boldsymbol{x}^{\top}\boldsymbol{A}^{-1}\boldsymbol{x}=\norm{\boldsymbol{x}}_{\boldsymbol{A}^{-1}}^2\label{pd}\,,
\end{align}
where Eq. (\ref{woodbury matrix identity}) follows from the Woodbury matrix identity \cite{woodbury1950inverting}, and Eq. (\ref{pd}) is because $(\boldsymbol{B}^{-1}+\boldsymbol{A}^{-1})^{-1}$ is a positive definite matrix.

With the above result, then following Eq. (\ref{equation13}) and the Cauchy–Schwarz inequality, we can get
\begin{equation}
        \begin{aligned}
                R(T)&\leq 2\sum_{t=1}^{T}C_{a_t,t}\\
               &=2\alpha_t\sum_{t=1} ^{T}\norm{\boldsymbol{x}_{a_t}}_{\boldsymbol{M}_t^{-1}}\\
               &\leq2\alpha_T\sum_{t=1} ^{T}\norm{\boldsymbol{x}_{a_t}}_{\boldsymbol{M}_t^{-1}}\\
                &\leq 2\alpha_T\sqrt{T\sum_{t=1}^{T}\norm{\boldsymbol{x}_{a_t}}^2_{\boldsymbol{M}^{-1}_t}}\\
                &\leq 2\alpha_T\sqrt{T\sum_{t=1}^{T}\norm{\boldsymbol{x}_{a_t}}^2_{\boldsymbol{V}^{-1}_t}}\,.
        \end{aligned}
        \label{equation14}
\end{equation}

Using Lemma 11 in \cite{abbasi2011improved}, with probability at least $1-\delta$, we can get
\begin{equation}
        \sum_{t=1}^T\norm{\boldsymbol{x}_{a,t}}_{\boldsymbol{V}^{-1}_t}^2
        \leq 2\log\bigg(\frac{det(\boldsymbol{V}_T)}{det(\beta\boldsymbol{I})}\bigg)
        \label{equation15}\,.
\end{equation}

Following similar steps as in Eq. (\ref{trace equality}), we can get that
\begin{equation}
    \frac{det(\boldsymbol{V}_T)}{det(\beta\boldsymbol{I})}\leq \big(\frac{T+\beta d}{\beta d}\big)^d\,.
\end{equation}

Therefore we have
\begin{equation}
   \sum_{t=1}^T\norm{\boldsymbol{x}_{a,t}}_{\boldsymbol{V}^{-1}_t}^2 \leq2d\log(1+\frac{T+1}{\beta d})\,. 
   \label{final}
\end{equation}

The result then follows by plugging in the definition of $\alpha_T$ and Eq. (\ref{final}) into Eq. (\ref{equation14}).
\end{proof}




\end{document}

%% file: Acknowledgement.tex
\section{Acknowledgement}
The corresponding author Shuai Li is supported by National Natural Science Foundation of China (62006151) and Shanghai Sailing Program. The work of John C.S. Lui was supported in part by the RGC's GRF 14215722.